\documentclass[sigconf, nonacm]{acmart}
\AtBeginDocument{%
  }

\setcopyright{acmlicensed}
\copyrightyear{2026}
\acmYear{2026}
\acmDOI{XXXXXXX.XXXXXXX}
\acmConference[Conference acronym 'XX]{Make sure to enter the correct conference title from your rights confirmation email}{June 03--05, 2018}{Woodstock, NY}
\acmISBN{978-1-4503-XXXX-X/XXXX/XX}

\usepackage[ruled,vlined,linesnumbered]{algorithm2e}

\newcommand{\E}{\mathcal{E}}
\newcommand{\R}{\mathcal{R}}
\newcommand{\F}{\mathcal{F}}

\newcommand{\U}{\mathcal{U}}
\newcommand{\D}{\mathcal{D}}
\newcommand{\pospart}[1]{\left[#1\right]_{+}}

\SetKwInput{AlgoInput}{Input}
\SetKwInput{AlgoOutput}{Output}

\newcounter{ToDo}
\newcounter{gaocomm} 
\newcounter{Note}
\definecolor{blue-violet}{rgb}{0.00,0.75,0.90}
\definecolor{mygreen}{rgb}{0.0, 0.5, 0.0}
\definecolor{awesome}{rgb}{1.0, 0.13, 0.32}
\definecolor{bostonuniversityred}{rgb}{1.0, 0.0, 0.0}
\begin{document}

\title{Matched Excess-Outranker Regularization for Candidate-Set Interference in Continual Knowledge Graph Embedding}


\author{Hao Ren}
\affiliation{%
  \institution{University of New South Wales}
  \city{Sydney}
  \state{NSW}
  \country{Australia}
}
\email{hao.ren@unsw.edu.au}
\orcid{0000-0003-2169-0111}

\author{Junbin Gao}
\affiliation{%
  \institution{The University of Sydney}
  \city{Sydney}
  \state{NSW}
  \country{Australia}
}
\email{junbin.gao@sydney.edu.au}
\orcid{0000-0001-9803-0256}

\author{Jiaojiao Jiang}
\correspondingauthor
\affiliation{%
  \institution{University of New South Wales}
  \city{Sydney}
  \state{NSW}
  \country{Australia}
}
\email{jiaojiao.jiang@unsw.edu.au}
\orcid{0000-0001-7307-8114}

\renewcommand{\shortauthors}{Ren et al.}

\begin{abstract}
Continual knowledge graph embedding updates entity and relation representations as a graph grows. Existing methods primarily address catastrophic forgetting, but entity admission also changes the candidate universe of every compatible query. A historical answer can therefore lose rank even when its score and its ordering among old entities are preserved. We formalize this effect as candidate-set interference and introduce Matched Excess-Outranker Regularization (\textsc{Meor}), a host-level objective that compares smooth answer-relative newcomer pressure with score-blind, structurally matched old references. Its one-sided penalty acts only when newcomer competition exceeds the matched reference, preserving the host learner's signal for legitimate new entities. Across eight paired runs on ENTITY--ComplEx, \textsc{Meor} improves historical current-universe mean reciprocal rank (MRR) by $0.0057$ over replay and reduces candidate-set interference by $0.0055$, with one-sided 95\% lower bounds of $0.0052$ and $0.0051$, respectively. It satisfies the preservation criteria for old-universe ranking and newcomer acquisition and improves historical current-universe MRR over persistent calibration, matched maximum regularizer (MMR), and unmatched old regularizer (UOR). Direct ablations support each component of its reference construction and aggregation. Adding \textsc{Meor} also improves historical ranking in all ten reported FBInc-S and FBInc-L host and backbone settings, with every paired 95\% confidence interval excluding zero. These results establish candidate admission as a distinct source of continual rank loss and show that it can be controlled without replacing the underlying embedding architecture or continual learner.
\end{abstract}

%%
%% ACM Computing Classification System
%% https://dl.acm.org/ccs
%%




\maketitle

\section{Introduction}

Knowledge graphs (KGs) in deployed systems expand as new entities and facts become available, while link prediction must remain reliable for knowledge already represented \cite{2022PredictionKGE,2024KGE}. Continual knowledge graph embedding (KGE) supports this process by updating a model without retraining it from the beginning. Existing continual KGE research has focused primarily on catastrophic forgetting, in which learning from new facts changes representations acquired earlier. This account is incomplete when graph growth expands the entity set. Every admitted entity joins the candidate universe of each compatible query and can displace an established answer even when the answer score and its ordering relative to every previously known entity remain unchanged \cite{2026EntityInterference}. The research problem addressed in this paper is how to control this admission-induced historical rank loss without materially weakening the model's ability to learn newly admitted entities when they are correct answers. We address this problem with Matched Excess-Outranker Regularization (\textsc{Meor}), a host-level objective that suppresses only the newcomer pressure exceeding a score-blind, structurally matched old-candidate reference.

\begin{figure}[t]
\centering
\includegraphics[width=\linewidth]{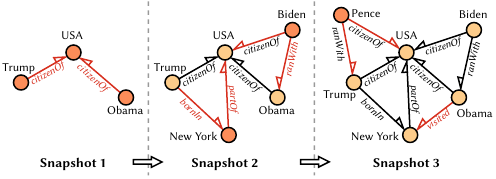}
\caption{Cumulative evolution of a continual knowledge graph across three snapshots.}
\Description{Dark orange nodes and red edges mark new knowledge, while light orange nodes and black edges show retained knowledge.}
\label{fig}
\end{figure}

Figure~\ref{fig} illustrates the cumulative entity admission that creates this additional competition. We study the resulting failure mode as candidate-set interference. Unlike catastrophic forgetting, it does not require a change in model parameters or in the relative ordering of old entities. It can be isolated at a fixed checkpoint by evaluating the same historical query over two candidate universes. The old universe measures ranking among previously known entities, whereas the current universe additionally contains the entities admitted through graph growth. Their difference identifies the rank loss attributable to newcomer competition. Current-universe evaluation exposes this effect, but it does not provide a training rule for controlling it. Replay, distillation, and parameter regularization protect previously learned representations, yet they do not directly regulate the query-specific competition created when new entities enter the candidate universe \cite{2020DiCGRL,2021ContinualLearningKGE,2023LKGE,2024IncDE}. Recent work identifies entity interference as an evaluation concern \cite{2026EntityInterference}. Our central contribution is to convert this diagnostic distinction into an explicit objective for continual learning.

\textsc{Meor} operates on historical replay queries during continual refinement \cite{2019Replay}. For each query, it measures the positive score gaps between newly admitted candidates and the recorded answer, normalizes these gaps to the local score scale, and smoothly aggregates the above-answer tail. It then compares this newcomer pressure with pressure from old candidates matched by prediction role, degree, and incident relation-role structure. The matching procedure does not use model scores, so the reference cannot adapt to the outcome against which it is evaluated. A one-sided squared hinge penalizes only the pressure exceeding the matched reference. Newcomers whose aggregate pressure remains consistent with structurally comparable old candidates receive no penalty, while the original host objective continues to govern the learning of new facts. \textsc{Meor} therefore targets the harmful component of admission-induced competition without replacing the embedding architecture or continual update rule of the host.

The experiments support both the ranking benefit and the proposed mechanism. Across eight paired runs on ENTITY--ComplEx, \textsc{Meor} improves historical current-universe mean reciprocal rank (MRR) by $0.0057$ over replay, with a one-sided 95\% lower bound of $0.0052$, and reduces candidate-set interference by $0.0055$, with a lower bound of $0.0051$. Relative to replay, the mean changes are $0.0002$ in old-universe MRR and $0.0015$ in target-newcomer MRR, and both lower bounds satisfy the preservation rule. \textsc{Meor} also improves historical ranking over persistent calibration, Matched Maximum Regularizer (MMR), and Unmatched Old Regularizer (UOR). Direct ablations show positive contributions from smooth tail aggregation, structural matching, query-specific assignment, and old-reference centering. On FBInc-S and FBInc-L, adding \textsc{Meor} improves historical current-universe MRR in all ten reported combinations of stream, backbone, and continual host. Every paired 95\% confidence interval excludes zero, and every newcomer-acquisition bound remains within the preservation margin.

This work makes three contributions. First, we formalize candidate-set interference as a continual KGE optimization problem distinct from catastrophic forgetting and provide a same-checkpoint rank decomposition that separates admission-induced competition from changes in the ordering of old entities. Second, we introduce \textsc{Meor}, a differentiable intervention that combines answer-relative smooth tail aggregation, score-blind structural matching, and one-sided excess regularization while preserving the host learning objective. Third, we provide a paired, mechanism-focused empirical analysis that connects historical-ranking gains to reduced candidate-set interference, isolates the contribution of each design component, and demonstrates transfer across the reported graph streams, embedding backbones, and continual-learning hosts.

\section{Related Work}

\subsection{Static KGE and Ranked Evaluation}

Knowledge graph embedding represents entities and relations in a continuous space and assigns a plausibility score to each candidate triple. Translation-based models such as TransE interpret a relation as a displacement between entity embeddings \cite{2013TransE}. Bilinear models score triples through multiplicative interactions, and ComplEx extends this family to complex-valued embeddings, allowing a single scoring form to represent both symmetric and antisymmetric relations \cite{2016ComplEx}. More recent probabilistic approaches, such as Normalizing Flow Embedding, replace deterministic point embeddings with distributional representations and score triples through similarities between learned flows, providing a richer treatment of representation uncertainty \cite{2023NFE}. These models differ in geometry and scoring function, but they share the same static link-prediction interface: given a query formed by masking one endpoint of a triple, the model scores candidate entities and ranks them by plausibility.

Although KGE models differ in representation geometry, they are commonly assessed through link prediction, where learned triple scores are converted into ranked candidate lists. A link-prediction query masks the head or tail entity of a known fact, scores the eligible replacement entities, and ranks its recorded answer among them. Filtered evaluation removes other known correct answers from the candidate list so that valid alternatives are not counted as errors \cite{2021LinkPrediction}. The resulting rank therefore depends not only on the learned score function, but also on the filtering rule and the candidate universe. In static benchmarks, this universe is usually fixed. In a growing knowledge graph, however, newly admitted entities become additional competitors for historical queries. Candidate construction therefore becomes part of the continual learning problem, rather than a neutral evaluation detail.

\subsection{Temporal and Continual KGE}

Temporal KGE and continual KGE address different forms of change. Temporal KGE treats time as part of the fact being modeled: facts are associated with timestamps or validity intervals, and these temporal annotations directly condition the learned representations or scoring function \cite{2024Temporal}. HyTE represents each timestamp as a hyperplane on which entities and relations are projected \cite{2018HyTE}, while diachronic embedding makes entity representations explicit functions of time \cite{2020Diachronic}. In these models, time is semantic information used at prediction time.

Continual KGE addresses a different setting: the model is updated as facts, entities, or relations are admitted across successive graph states. The update order determines what information is available at each stage, but time need not be a semantic argument of the score function. A graph state may reflect data acquisition, benchmark construction, or incremental deployment rather than a factual validity interval. Thus, temporal KGE asks which fact holds at a specified time, whereas continual KGE asks how an existing representation can be preserved and extended as the graph  \cite{2026Semantic}.

Continual KGE treats graph evolution as an incremental representation learning problem and adapts continual-learning strategies to update knowledge graph embeddings without relearning all concepts from scratch \cite{2021ContinualLearningKGE}. Representative methods use selected historical data and disentangled components \cite{2020DiCGRL, 2021ContinualLearningKGE}, masked reconstruction, embedding transfer, and distillation \cite{2023LKGE, 2024IncDE}, or efficient adaptation, dynamic regularization, coordinated masks, and task-conditioned transfer \cite{2024FastKGE, 2024FMR, 2025CMKGE, 2025ETT-CKGE}. Despite their architectural differences, these methods share the goal of retaining useful knowledge while incorporating each graph update.

\subsection{Catastrophic Forgetting in Continual KGE}

The central concern in continual KGE is catastrophic forgetting \cite{2023Forgetting}. Training on newly available facts can move established representations and reduce performance on historical knowledge, creating a tension between stability and plasticity. The model must preserve useful structure from earlier graph states while remaining responsive to new entities, relations, and facts.

Existing methods address this problem through complementary strategies. Replay retains selected historical observations, regularization limits destructive parameter change, and distillation transfers information from an earlier model state \cite{2019Replay}. Selective and modular updates restrict which representations are modified, while reconstruction objectives preserve information encoded in the earlier graph \cite{2020DiCGRL, 2021ContinualLearningKGE, 2023LKGE, 2024IncDE}. FMR combines rotated representations with dynamic regularization \cite{2024FMR}, CMKGE coordinates stability and plasticity through dual masks \cite{2025CMKGE}, and Dynamic Temperature Distillation adjusts knowledge transfer according to fact volatility \cite{2026DTD}. Other work improves continual updates through informed initialization \cite{2026FBinc}, adaptive representation capacity \cite{2025SAGE}, structure-aware representation learning \cite{2026STARK}, and Bayesian-guided evolution \cite{2026BAKE}.

These methods primarily control how learned representations change between graph states. Their evaluation therefore emphasizes whether knowledge encoded before an update remains recoverable afterward. This perspective captures parameter and representation forgetting, but it does not fully describe the ranking consequences of enlarging the entity universe.

\subsection{Candidate-Set Interference}

Candidate-set interference is a distinct source of historical rank loss. When new entities enter the graph, they become eligible answers to earlier queries. A historical answer can therefore move down in the ranking even when its score and its ordering relative to every old entity remain unchanged. The loss arises because the query faces additional competitors, not because the model has forgotten how to rank established entities.

Pons et al. \cite{2026EntityInterference} identify this effect as entity interference and show that evaluation restricted to old candidates overstates continual performance. A same-checkpoint comparison makes the distinction precise. Evaluating the same historical query first over the old candidate universe and then over the expanded current universe isolates the rank loss attributable to admitted candidates. Catastrophic forgetting instead concerns changes caused by model updating. Candidate-set interference can occur without forgetting, and forgetting can occur while the candidate universe remains fixed. The two mechanisms are complementary explanations for historical degradation.

Current-universe evaluation exposes candidate-set interference, but the continual objectives reviewed above do not directly regulate the ranking pressure that admitted entities create for each historical answer. This intervention gap defines the problem addressed here. We treat candidate competition as an explicit learning target within a fixed continual KGE process. The next section formalizes the resulting rank decomposition and evaluation objectives.

\section{Continual KGE Problem Formulation}
\label{sec:setting}

\subsection{Graph Growth and Candidate Admission}

Let $\mathcal{G}_0,\ldots,\mathcal{G}_T$ denote a sequence of cumulative knowledge-graph snapshots. At snapshot $u$, the updated graph is represented as $\mathcal{G}_u=(\E_u,\R_u,\F_u)$, where $\E_u$ and $\R_u$ denote the sets of entities and relations observed up to snapshot $u$, respectively, and $\F_u\subseteq \E_u\times\R_u\times\E_u$ denotes the corresponding set of observed training triples. The continual setting assumes
\begin{equation}
  \E_{u-1}\subseteq\E_u,\qquad
  \R_{u-1}\subseteq\R_u,\qquad
  \F_{u-1}\subseteq\F_u.
\end{equation}
The entities and relations admitted at update $u$ are
\begin{equation}
  \Delta\E_u=\E_u\setminus\E_{u-1},
  \qquad
  \Delta\R_u=\R_u\setminus\R_{u-1}.
\end{equation}

We write $\U\subseteq\{1,\ldots,T\}$ for the evaluated updates. For a fact $(h,r,t)$, tail prediction uses the query $q=(h,r,\cdot)$ with answer $a=t$, whereas head prediction uses $q=(\cdot,r,t)$ with answer $a=h$. We therefore let $\D\subseteq\{\mathrm{head},\mathrm{tail}\}$ denote the evaluated prediction modes and use $d\in\D$ to identify the missing endpoint. The notation $r(q)$ refers to the relation in $q$, and $e$ denotes a candidate entity. The experimental choices of $\U$ and $\D$ appear in Section~\ref{sec:experiments}.

At update $u$, a KGE model with parameters $\theta_{m,k,u}$ assigns score $s_{m,k,u}(q,e)$ to candidate $e$ for query $q$. The indices $m$ and $k$ identify the learning method and training realization. Higher scores indicate stronger preference. The formulation is independent of the embedding architecture.

\subsection{Historical Rank Loss Under Candidate Growth}

Consider a historical query occurrence $x=(q,a,u,d)$ whose source fact appeared before update $u$. Its answer $a$ belongs to $\E_{u-1}$. Using the same model state, let $C_x^{\mathrm{old}}$ and $C_x^{\mathrm{cur}}$ be the filtered candidate sets drawn from $\E_{u-1}$ and $\E_u$. The same filtering rule is applied to both sets, and the answer is retained. Consequently, $C_x^{\mathrm{old}}\subseteq C_x^{\mathrm{cur}}$. We write $\Delta C_x=C_x^{\mathrm{cur}}\setminus C_x^{\mathrm{old}}$ for the candidates introduced by the expansion.

The score function together with a fixed tie rule induces a total order. We write $e\succ_{m,k,x} a$ when candidate $e$ precedes answer $a$ under model state $(m,k,u)$. For $v\in\{\mathrm{old},\mathrm{cur}\}$, the corresponding rank is
\begin{equation}
  r^v_{m,k,u}(x)
  =1+\sum_{e\in C_x^v\setminus\{a\}}\mathbb{I}[e\succ_{m,k,x} a].
\end{equation}
The two ranks satisfy
\begin{equation}
  r^{\mathrm{cur}}_{m,k,u}(x)
  =r^{\mathrm{old}}_{m,k,u}(x)
  +\sum_{e\in\Delta C_x}\mathbb{I}[e\succ_{m,k,x} a].
  \label{eq:rank-decomposition}
\end{equation}
This identity isolates the rank loss attributable to newly admitted candidates. It holds even when the ordering among old entities is unchanged and makes no assumption about whether their representations also change. Evaluation in the current candidate universe therefore reveals a source of interference that evaluation over old candidates alone cannot measure \cite{2026EntityInterference}.

\subsection{Query Roles and Evaluation Objectives}

We distinguish three query roles. A historical occurrence comes from a fact observed before the current update. A target newcomer occurrence has an answer in $\Delta\E_u$. A query newcomer occurrence has a known query entity in $\Delta\E_u$ and an answer in $\E_{u-1}$. The latter roles capture different learning requirements and are therefore evaluated separately.

For data split $\sigma$, let $\mathcal{C}^{\sigma,\mathrm{H}}\subseteq\U\times\D$ be the predefined historical evaluation cells. Let $Q^{\sigma,\mathrm{H}}_{u,d}$ contain the historical occurrences in cell $(u,d)$. Historical MRR in the current candidate universe is
\begin{equation}
H^{\sigma}_{\mathrm{cur},m,k}
=\frac{1}{|\mathcal{C}^{\sigma,\mathrm{H}}|}
\sum_{(u,d)\in\mathcal{C}^{\sigma,\mathrm{H}}}
\frac{1}{|Q^{\sigma,\mathrm{H}}_{u,d}|}
\sum_{x\in Q^{\sigma,\mathrm{H}}_{u,d}}
\frac{1}{r^{\mathrm{cur}}_{m,k,u}(x)}.
\label{eq:hcur}
\end{equation}
Replacing current ranks with old candidate ranks defines $H^{\sigma}_{\mathrm{old},m,k}$. Their difference is
\begin{equation}
  D^{\sigma}_{\mathrm{MCI},m,k}
  =H^{\sigma}_{\mathrm{old},m,k}-H^{\sigma}_{\mathrm{cur},m,k}.
\label{eq:dmci}
\end{equation}
This quantity measures the loss associated with candidate expansion. A small value is not sufficient evidence of useful ranking because both terms can decrease together. We therefore treat current-universe historical MRR as the primary utility outcome and old-universe historical MRR as a preservation outcome.

Let $\mathcal{C}^{\sigma,\mathrm{TN}}\subseteq\U\times\D$ be the predefined target newcomer cells, and let $Q^{\sigma,\mathrm{TN}}_{u,d}$ be their query sets. Target newcomer acquisition is measured by
\begin{equation}
A^{\sigma}_{\mathrm{TN},m,k}
=\frac{1}{|\mathcal{C}^{\sigma,\mathrm{TN}}|}
\sum_{(u,d)\in\mathcal{C}^{\sigma,\mathrm{TN}}}
\frac{1}{|Q^{\sigma,\mathrm{TN}}_{u,d}|}
\sum_{x\in Q^{\sigma,\mathrm{TN}}_{u,d}}
\frac{1}{r^{\mathrm{cur}}_{m,k,u}(x)}.
\label{eq:atn}
\end{equation}
Each included cell receives equal weight, and queries receive equal weight within a cell. An endpoint is defined only when every predefined query set is nonempty. An empty set is never assigned zero. Query newcomer performance is constructed in the same manner. When the split, method, and training realization are clear, we abbreviate the four endpoints as $H_{\mathrm{cur}}$, $H_{\mathrm{old}}$, $D_{\mathrm{MCI}}$, and $A_{\mathrm{TN}}$.

\section{Matched Excess-Outranker Regularization}
\label{sec:method}

Equation~\eqref{eq:rank-decomposition} motivates an answer-relative training signal. Newly admitted entities matter when they move above a historical answer. A raw outranker count is discontinuous, while an unnormalized score gap is not comparable across queries. Penalizing newcomers against zero or an arbitrary old sample also confounds cohort identity with visible structural differences.

\textsc{Meor} combines a smooth, normalized measure of newcomer competition relative to the answer with an old reference matched on observable structure. It penalizes only the portion of newcomer competition that exceeds this reference. Consequently, a large newcomer aggregate incurs no penalty when a structurally comparable old cohort produces a similar value.

\begin{figure*}[t]
  \centering
  \includegraphics[width=\textwidth]{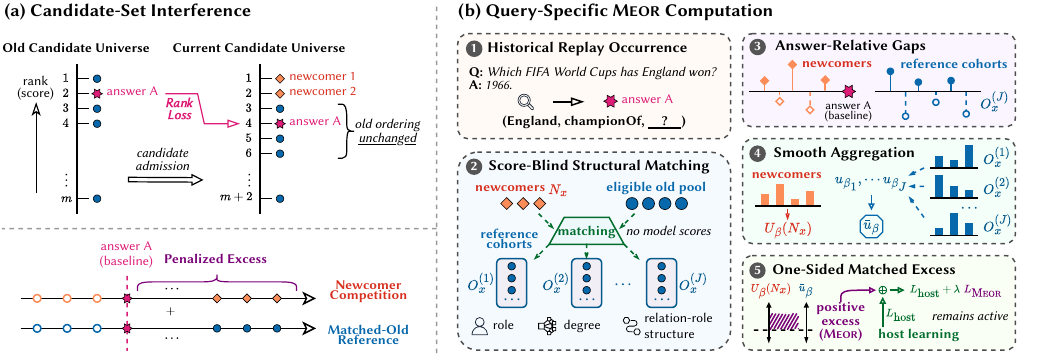}
  \caption{Candidate-set interference and the query-specific computation of \textsc{Meor}.}
  \Description{The upper-left panel compares a historical query under old and expanded candidate universes, where the answer loses rank while old candidates retain their relative order. The upper-right panel contrasts newcomer competition with a matched-old reference. The lower panel traces a replay occurrence through score-blind structural matching, answer-relative gaps, smooth aggregation, and the one-sided \textsc{Meor} penalty combined with the host loss.}
  \label{fig:meor}
\end{figure*}

Figure~\ref{fig:meor} illustrates how candidate-set interference arises and how \textsc{Meor} addresses it.

\subsection{Answer-Relative Newcomer Competition}

The regularizer operates on historical occurrences drawn from replay. Throughout this section, we fix a method $m$ and training realization $k$; every score-derived quantity below inherits these indices. For an occurrence $x=(q,a,u,d)$, let $P^{\mathrm{tr}}_u(q,d)$ contain the answers visible in cumulative training facts through update $u$. The eligible newcomer cohort is
\begin{equation}
N_x = \operatorname{sort}_{\mathrm{id}}\!\left(
\Delta\E_u\setminus\bigl(\{a\}\cup P^{\mathrm{tr}}_u(q,d)\bigr)
\right).
\label{eq:new-cohort}
\end{equation}
The complete eligible old pool is
\begin{equation}
\E^{-,\mathrm{tr}}_{u-1}(x)=\operatorname{sort}_{\mathrm{id}}\!\left(
\E_{u-1}\setminus\bigl(\{a\}\cup P^{\mathrm{tr}}_u(q,d)\bigr)
\right).
\label{eq:old-pool}
\end{equation}
Both sets are defined only from the training prefix. Validation and test labels therefore cannot affect cohort membership. The answer is excluded explicitly, even when it is absent from $P^{\mathrm{tr}}_u(q,d)$.

We draw a scale sample $C_x^{\mathrm{scale}}$ from the eligible old pool without using model scores, with
\begin{equation}
|C_x^{\mathrm{scale}}|=\min\left\{K_{\mathrm{scale}},|\E^{-,\mathrm{tr}}_{u-1}(x)|\right\}.
\end{equation}
Its identifiers are selected independently of model scores and shared by paired methods. This prevents the observed scores from influencing normalization.

Let $\operatorname{IQR}_x$ be the interquartile range of the detached finite scores in $C_x^{\mathrm{scale}}$ \cite{2005IQR}. Independently, let $\operatorname{MAD}_x^{\mathrm{fb}}$ be the median absolute deviation from the first adequate training-prefix pool in the following hierarchy: the same relation and prediction mode, the same prediction mode, and the complete update. We set this term to zero when no pool is adequate and define
\begin{equation}
\tau_x=\max\left\{\operatorname{IQR}_x,
\eta_{\mathrm{MAD}}\operatorname{MAD}_x^{\mathrm{fb}},\varepsilon_\tau\right\}.
\label{eq:scale}
\end{equation}
The scale is detached from the gradient. It makes local score differences dimensionless and limits the influence of queries with unusually large score ranges. The parameters $\eta_{\mathrm{MAD}}>0$ and $\varepsilon_\tau>0$ are numerical safeguards rather than learned quantities. The pooled fallback is robust to isolated extreme scores but can reflect shifts in score location across occurrences. It is therefore not solely a measure of dispersion within individual queries.

For candidate $e$, the normalized gap relative to the answer is
\begin{equation}
z_x(e)=\frac{s_{m,k,u}(q,e)-s_{m,k,u}(q,a)}{\tau_x}.
\label{eq:z}
\end{equation}

For a nonempty ordered candidate multiset $C=(e_1,\ldots,e_n)$, define
\begin{equation}
U_{\beta}(x,C)=\frac{1}{\beta}
\log\left(
\frac{1}{n}\sum_{i=1}^{n}
\exp\left\{\beta\pospart{z_x(e_i)}\right\}
\right), \qquad \beta>0.
\label{eq:aggregate}
\end{equation}
The normalization by $n$ prevents cohort size alone from increasing the aggregate and makes it invariant to exact replication of the complete multiset. A repeated entity remains a repeated score instance. The aggregate is zero exactly when no member scores above the answer, and it never exceeds the largest positive normalized gap in $C$. The parameter $\beta$ determines how strongly the largest active gaps influence the smooth aggregate. The quantity is neither an outranker count nor a probability.

\subsection{Score-Blind Matched References}

Inspired by Jones and Love \cite{2007Role}, we assume old reference should resemble the newcomer cohort in observable structural characteristics while remaining independent of the scores being compared. For entity $e$ and prediction mode $d$, let $b_u(e,d)$ be its endpoint-degree bin and let $\rho_u(e,d)$ be its set of incident relation-role pairs, both computed from the cumulative training prefix. Matching uses three nested, score-blind keys. The finest key retains $(d,b_u(e,d),\rho_u(e,d))$, the next drops the relation-role signature, and the terminal key retains only $d$. For each newcomer, we select the first key whose corresponding old-entity cell is nonempty. Each coarser key only merges cells formed at the preceding level.

Within the selected cell, sampling proceeds without replacement until the cell is exhausted and uses replacement only for any remaining positions. Concatenating the selected candidates across newcomer instances yields an ordered old multiset $O_x^{(j)}$ with $|O_x^{(j)}|=|N_x|$.

We construct $J$ matched draws. Their candidate identifiers are fixed independently of scores and shared by paired methods. Each model then supplies its own scores for those candidates. This design reduces observable cohort imbalance without treating the matched references as causal counterfactuals.

\subsection{Matched Excess Objective}

Let $\mathcal{B}_u$ denote the current refinement batch. The ordered multiset $B_M$ contains its eligible replay query occurrences in batch order. An occurrence enters $B_M$ only when its newcomer cohort, scale sample, and terminal old matching pool are nonempty and every required score and scale input is finite. Membership is frozen before computing any method's aggregate and is shared by all paired controls. Repeated occurrences remain repeated instances. We write $\mathcal{L}_{\mathrm{host}}(\theta_{m,k,u};\mathcal{B}_u)$ for the differentiable refinement objective supplied by the underlying continual KGE learner before \textsc{Meor} is added. This objective retains the learner's positive training signal for current facts and any replay facts selected by the host.

The mean old reference is
\begin{equation}
\overline U_O(x)=\frac{1}{J}\sum_{j=1}^{J}
U_\beta\!\left(x,O_x^{(j)}\right).
\label{eq:old-reference}
\end{equation}
For nonempty $B_M$, the \textsc{Meor} loss is
\begin{equation}
\mathcal{L}_{\textsc{Meor}}=
\frac{1}{|B_M|}\sum_{x\in B_M}
\pospart{U_\beta(x,N_x)-\overline U_O(x)}^2.
\label{eq:meor}
\end{equation}
The mean in Equation~\eqref{eq:old-reference} is computed before applying the hinge to positive excess. If the newcomer aggregate equals the mean reference, its contribution is zero even when it exceeds an individual draw. Applying a separate hinge to each draw before averaging would impose downward pressure in that case. The squared hinge has both value and gradient zero at the boundary, and its penalty increases with positive excess.

If $B_M$ is empty, we define $\mathcal{L}_{\textsc{Meor}}=0$. With regularization coefficient $\lambda\geq 0$, the complete refinement objective is
\begin{equation}
\mathcal{L}(\theta_{m,k,u};\mathcal{B}_u)
=\mathcal{L}_{\mathrm{host}}(\theta_{m,k,u};\mathcal{B}_u)
+\lambda\mathcal{L}_{\textsc{Meor}}.
\label{eq:full-objective}
\end{equation}
The host objective retains its positive learning signal for newly admitted entities. Thus \textsc{Meor} controls excessive competition on historical queries rather than treating every newcomer as an error.

\begin{algorithm}[t]
  \AlgoInput{training-prefix graph at update $u$, refinement batch
  $\mathcal{B}_u$, model scores $s_{m,k,u}$, host objective
  $\mathcal{L}_{\mathrm{host}}$, and parameters $(J,\beta,\lambda)$}
  \AlgoOutput{regularized refinement objective
  $\mathcal{L}(\theta_{m,k,u};\mathcal{B}_u)$}

  $B_M\leftarrow\emptyset$\;

  \ForEach{historical replay occurrence $x=(q,a,u,d)\in\mathcal{B}_u$}{
    construct $N_x$ using Equation~\eqref{eq:new-cohort}\;
    draw the score-blind scale sample $C_x^{\mathrm{scale}}$\;
    construct $J$ score-blind matched cohorts
    $\{O_x^{(j)}\}_{j=1}^{J}$ with
    $|O_x^{(j)}|=|N_x|$\;

    \If{$x$ has all required nonempty cohorts and finite score and scale inputs}{
      $B_M\leftarrow B_M\uplus\{x\}$\;
    }
  }

  $R\leftarrow 0$\;

  \ForEach{$x\in B_M$}{
    compute the detached scale $\tau_x$ using
    Equation~\eqref{eq:scale}\;
    compute $z_x(e)$ using Equation~\eqref{eq:z} for all
    $e\in N_x\uplus\biguplus_{j=1}^{J}O_x^{(j)}$\;
    $\overline U_O(x)\leftarrow
    \dfrac{1}{J}\sum_{j=1}^{J}
    U_\beta\!\left(x,O_x^{(j)}\right)$\;
    $R\leftarrow R+
    \pospart{U_\beta(x,N_x)-\overline U_O(x)}^2$\;
  }

  $\mathcal{L}_{\textsc{Meor}}\leftarrow
  \begin{cases}
    0, & B_M=\emptyset,\\
    R/|B_M|, & \text{otherwise},
  \end{cases}$\;

  $\mathcal{L}(\theta_{m,k,u};\mathcal{B}_u)\leftarrow
  \mathcal{L}_{\mathrm{host}}(\theta_{m,k,u};\mathcal{B}_u)
  +\lambda\mathcal{L}_{\textsc{Meor}}$\;

  \caption{\textsc{Meor} objective for a refinement batch}
  \label{alg:meor}
\end{algorithm}

\subsection{Controls for the Proposed Mechanism}

The central contribution of \textsc{Meor} is to combine smooth answer-relative aggregation with score-blind structural matching. Two controls isolate these components. Both control losses are defined as zero when $B_M$ is empty; the displayed averages apply otherwise. The Matched Maximum Regularizer (MMR) replaces the smooth aggregate with
\begin{equation}
V(x,C)=\max_{1\leq i\leq |C|}\pospart{z_x(e_i)}
\end{equation}
and uses the same matched reference cohorts. Its loss is
\begin{equation}
\mathcal{L}_{\mathrm{MMR}}=\frac{1}{|B_M|}\sum_{x\in B_M}
\pospart{V(x,N_x)-\frac{1}{J}\sum_{j=1}^{J}V(x,O_x^{(j)})}^{2}.
\label{eq:mmr}
\end{equation}
This comparison isolates the effect of distributing pressure across the active score tail rather than controlling only the largest gap.

The Unmatched Old Regularizer (UOR) retains $U_\beta$ but replaces each matched reference with an equal-cardinality multiset $R_x^{(j)}$ drawn from the terminal role-only old pool. It uses
\begin{equation}
\mathcal{L}_{\mathrm{UOR}}=\frac{1}{|B_M|}\sum_{x\in B_M}
\pospart{U_\beta(x,N_x)-\frac{1}{J}\sum_{j=1}^{J}
U_\beta(x,R_x^{(j)})}^{2}.
\label{eq:uor}
\end{equation}
This comparison isolates the contribution of structural matching. Replay uses the same host refinement with zero regularization and therefore isolates the contribution of the intervention.

A persistent admission-cohort calibrator provides a score-only control. Let $c(e)$ denote the update at which entity $e$ first entered the graph. For training realization $k$, $\alpha_{k,c,r,d}>0$ is a multiplicative coefficient and $g_{k,c,r,d}\in\mathbb{R}$ is a dimensionless offset coefficient. Both are fitted from replay validation scores for admission cohort $c$, relation $r$, and prediction mode $d$ when the cohort is admitted, then retained at later updates. The detached statistic $\bar{\tau}^{\mathrm{val}}_{k,u,r,d}$ is the median finite validation scale for the same relation and prediction mode at update $u$, with prediction-mode and update-level fallbacks. When this scale is available, the calibrator multiplies the raw replay score $s_{\mathrm{Replay},k,u}(q,e)$ by $\alpha_{k,c(e),r(q),d}$ and adds the scale-adjusted offset $g_{k,c(e),r(q),d}\bar{\tau}^{\mathrm{val}}_{k,u,r(q),d}$. If no finite scale summary is available, the raw replay score is retained. The initial cohort uses the identity transform, $\alpha=1$ and $g=0$. This control adds no training trajectory and separates an intervention applied during training from a post hoc adjustment of cohort score scales.

All controls share the host model, update data, replay information, optimization budget, and trainable parameter scope. The matched maximum control and \textsc{Meor} also share their matched reference identifiers. Their concrete training and calibration settings appear in Section~\ref{sec:experiments}.

\subsection{Interpretation and Computational Cost}

Equation~\eqref{eq:meor} has zero gradient whenever newcomer competition is no greater than the mean matched reference. When active, the smooth aggregate distributes pressure across several above-answer candidates rather than only the largest one. The loss acts on scores, so the evaluation measures historical current-universe ranking and target-newcomer acquisition directly. These outcomes capture whether score changes cross the answer and whether legitimate newcomer learning is preserved.

If one candidate score requires $O(p)$ operations, candidate scoring for one refinement batch without score reuse is
\begin{equation}
O\!\left(
p\sum_{x\in B_M}\bigl((1+J)|N_x|+K_{\mathrm{scale}}\bigr)
\right).
\label{eq:complexity}
\end{equation}
The complete newcomer cohort and every reference cohort are scored, so computational cost increases with the size of the admission batch. Approximate cohort scoring lies outside the scope of this study.

\section{Experiments and Results}
\label{sec:experiments}

The evaluation centers on whether \textsc{Meor} improves historical ranking by reducing candidate-set interference. ENTITY--ComplEx provides the primary comparison with replay, persistent calibration, MMR, and UOR. Direct ablations identify the contribution of each component, while experiments on FBInc-S and FBInc-L examine transfer across different rates of entity admission, embedding models, and continual KGE hosts. All comparative claims are drawn from paired runs within the same experimental setting.

\subsection{Experimental Setting}

Our experiments use four five-snapshot entity-growth streams. ENTITY is the primary FB15K-237 \cite{2015FB15k} stream introduced with LKGE \cite{2023LKGE}. FBInc-S and FBInc-L are the small-growth and large-growth streams distributed with the informed-initialization benchmark \cite{2026FBinc}. WN-CKGE is the WordNet stream used by FastKGE \cite{2024FastKGE}. It represents the applicability boundary discussed in Section~6 and is not treated as a positive efficacy setting. Snapshot~0 initializes the model, and Snapshots~1 through~4 form the continual sequence. Table~\ref{tab:planned-scope-premise} reports the cumulative graph size and the facts introduced at each snapshot.

\begin{table*}[t]
  \caption{Dataset statistics. Entity and relation counts are cumulative; Facts is the number of triples introduced at each snapshot.}
  \label{tab:planned-scope-premise}
  \centering
  \setlength{\tabcolsep}{4pt}
  \renewcommand{\arraystretch}{1.12}
  \begin{tabular}{@{}l*{5}{rrr}@{}}
    \toprule
    Dataset & \multicolumn{3}{c}{Snapshot 0} &
      \multicolumn{3}{c}{Snapshot 1} &
      \multicolumn{3}{c}{Snapshot 2} &
      \multicolumn{3}{c}{Snapshot 3} &
      \multicolumn{3}{c}{Snapshot 4} \\
    \cmidrule(lr){2-4}\cmidrule(lr){5-7}\cmidrule(lr){8-10}
    \cmidrule(lr){11-13}\cmidrule(l){14-16}
    & $|\mathcal{E}|$ & $|\mathcal{R}|$ & Facts
    & $|\mathcal{E}|$ & $|\mathcal{R}|$ & Facts
    & $|\mathcal{E}|$ & $|\mathcal{R}|$ & Facts
    & $|\mathcal{E}|$ & $|\mathcal{R}|$ & Facts
    & $|\mathcal{E}|$ & $|\mathcal{R}|$ & Facts \\
    \midrule
    ENTITY
      & $2{,}909$ & $233$ & $46{,}388$
      & $5{,}817$ & $236$ & $72{,}111$
      & $8{,}725$ & $236$ & $73{,}785$
      & $11{,}633$ & $237$ & $70{,}506$
      & $14{,}541$ & $237$ & $47{,}326$ \\
    FBInc-S
      & $2{,}909$ & $233$ & $46{,}388$
      & $2{,}919$ & $233$ & $235$
      & $2{,}930$ & $233$ & $152$
      & $2{,}940$ & $233$ & $180$
      & $2{,}950$ & $233$ & $239$ \\
    FBInc-L
      & $2{,}909$ & $233$ & $46{,}388$
      & $3{,}010$ & $233$ & $3{,}110$
      & $3{,}110$ & $234$ & $2{,}736$
      & $3{,}211$ & $234$ & $2{,}908$
      & $3{,}312$ & $234$ & $3{,}105$ \\
    WN-CKGE
      & $24{,}567$ & $11$ & $55{,}801$
      & $28{,}660$ & $11$ & $9{,}300$
      & $32{,}754$ & $11$ & $9{,}300$
      & $36{,}848$ & $11$ & $9{,}300$
      & $40{,}943$ & $11$ & $9{,}302$ \\
    \bottomrule
  \end{tabular}
\end{table*}

We evaluate filtered head and tail prediction after every update. The filter removes other positives visible in the current graph prefix while retaining the recorded answer, and exact score ties are resolved by increasing canonical entity identifier. Each endpoint assigns equal weight to the update and prediction-direction combinations defined in Section~\ref{sec:setting}. Historical current-universe MRR $H_{\mathrm{cur}}$ measures the ranking of established answers against every entity available at the current update. Historical old-universe MRR $H_{\mathrm{old}}$ restricts the same evaluation to previously known candidates. Their difference, $D_{\mathrm{MCI}}=H_{\mathrm{old}}-H_{\mathrm{cur}}$, measures the rank loss caused by candidate admission. Target-newcomer MRR $A_{\mathrm{TN}}$ evaluates queries whose correct answer is newly admitted.

Although \textsc{Meor} is independent of the embedding architecture, the primary study uses ComplEx with 200 complex coordinates and float64 arithmetic. Every method optimizes the same softplus link-prediction objective with Adam, a learning rate of $10^{-4}$, $(\beta_1,\beta_2)=(0.9,0.999)$, $\epsilon=10^{-8}$, and no weight decay. Training uses batches of $2{,}048$ positive facts with ten negative samples per fact. Base training runs for at most 200 epochs with an early-stopping patience of three, and replay retains at most $2{,}048$ historical facts. Continual refinement makes one ordered pass over the facts introduced at the current snapshot. Previously learned entity and relation embeddings remain fixed during refinement, so the intervention updates only newly admitted parameters.

We instantiate \textsc{Meor} with $\beta=5$, four matched reference cohorts, and at most 256 old entities in each scale sample. Reference construction matches prediction role, degree bin, and incident relation-role signature before falling back to role and degree and then to role alone. The regularization coefficient for each ENTITY method is fixed before evaluation by the score-blind gradient-balance rule. Replay uses zero regularization. Persistent calibration applies the affine transformation $\alpha s+g\bar{\tau}^{\mathrm{val}}$ to newcomer scores over the fixed grid
\begin{equation*}
\alpha\in\{0.5,0.75,1,1.25,1.5\},
\qquad
g\in\{-1,-0.5,0,0.5,1\},
\end{equation*}
where $\bar{\tau}^{\mathrm{val}}$ is the corresponding replay-validation score scale. Calibration retains transformations whose $A_{\mathrm{TN}}$ is no more than $0.005$ below replay and selects the retained transformation with the highest $H_{\mathrm{cur}}$.

The primary, ablation, and transfer studies each use eight paired runs. Within a pair, the compared methods share initialization, graph stream, fact order, replay sample, negative samples, stopping decisions, and evaluation queries. Comparisons involving matched references also share the relevant reference identities. The direct ablations use the same coefficient-selection rule and change only the component named in Table~\ref{tab:entity-direct-ablation}. The transfer study adds \textsc{Meor} to replay with ComplEx \cite{2016ComplEx}, DistMult \cite{2015DisMult}, and TransE \cite{2013TransE} and to the LKGE \cite{2023LKGE} and IncDE \cite{2024IncDE} update procedures with their TransE backbones. Comparisons are made only within the same dataset, host, and backbone.

For the primary and ablation studies, LB denotes the one-sided 95\% lower confidence bound computed from the eight unrounded paired effects. A positive lower bound supports an improvement in $H_{\mathrm{cur}}$ or a reduction in $D_{\mathrm{MCI}}$. Preservation requires the lower bound for $\Delta H_{\mathrm{old}}$ or $\Delta A_{\mathrm{TN}}$ to remain above $-0.005$. W/T/L counts positive, zero, and negative full-precision $H_{\mathrm{cur}}$ effects. The transfer study reports two-sided paired 95\% confidence intervals for $H_{\mathrm{cur}}$ and the same one-sided lower bound for $\Delta A_{\mathrm{TN}}$. These intervals are unadjusted summaries of prespecified, cell-specific comparisons rather than simultaneous family-wise intervals.

\subsection{Primary Results on ENTITY}

Table~\ref{tab:entity-absolute} reports the endpoint means for ENTITY with ComplEx. At $0.0887$, \textsc{Meor} attains the highest $H_{\mathrm{cur}}$. It also records the highest $H_{\mathrm{old}}$ and the lowest $D_{\mathrm{MCI}}$, while increasing $A_{\mathrm{TN}}$ above replay, MMR, and UOR. Persistent calibration attains the highest $A_{\mathrm{TN}}$ but a lower $H_{\mathrm{cur}}$. The calibration analysis reports only $H_{\mathrm{cur}}$ and $A_{\mathrm{TN}}$; $H_{\mathrm{old}}$ and $D_{\mathrm{MCI}}$ were not recorded for this control. All endpoints are equal-weight means over update and prediction-direction combinations.

\begin{table}[t]
  \caption{Absolute endpoint performance of \textsc{Meor} and mechanism-matched controls in the primary ENTITY--ComplEx setting.}
  \label{tab:entity-absolute}
  \centering
  \setlength{\tabcolsep}{4pt}
  \renewcommand{\arraystretch}{1.08}
  \begin{tabular}{@{}lcccc@{}}
    \toprule
    Method & $H_{\mathrm{cur}}\uparrow$ & $H_{\mathrm{old}}\uparrow$ &
      $D_{\mathrm{MCI}}\downarrow$ & $A_{\mathrm{TN}}\uparrow$ \\
    \midrule
    Replay & $0.0830$ & $0.1199$ & $0.0369$ & $0.1700$ \\
    Persistent calibration & $0.0816$ & -- & -- & $0.1753$ \\
    MMR & $0.0836$ & $0.1200$ & $0.0364$ & $0.1695$ \\
    UOR & $0.0842$ & $0.1200$ & $0.0358$ & $0.1698$ \\
    \textsc{Meor} & $0.0887$ & $0.1201$ & $0.0314$ & $0.1715$ \\
    \bottomrule
  \end{tabular}
\end{table}

The absolute endpoints show the aggregate ordering, while the paired analysis determines whether that ordering is consistent across runs. Table~\ref{tab:entity-paired-effects} reports the paired effects of \textsc{Meor} relative to each mechanism-matched control.

\begin{table*}[t]
  \caption{Paired effects of \textsc{Meor} relative to mechanism-matched controls on ENTITY--ComplEx across eight paired training realizations.}
  \label{tab:entity-paired-effects}
  \centering
  \setlength{\tabcolsep}{4pt}
  \renewcommand{\arraystretch}{1.10}
  \begin{tabular}{@{}lccccccccc@{}}
    \toprule
    & \multicolumn{2}{c}{$\Delta H_{\mathrm{cur}}\uparrow$}
    & \multicolumn{2}{c}{$\Delta H_{\mathrm{old}}\uparrow$}
    & \multicolumn{2}{c}{$D_{\mathrm{MCI}}$ reduction$\uparrow$}
    & \multicolumn{2}{c}{$\Delta A_{\mathrm{TN}}\uparrow$}
    & \\
    \cmidrule(lr){2-3}
    \cmidrule(lr){4-5}
    \cmidrule(lr){6-7}
    \cmidrule(lr){8-9}
    Comparator
    & Mean & LB
    & Mean & LB
    & Mean & LB
    & Mean & LB
    & W/T/L \\
    \midrule
    Replay
    & $0.0057$ & $0.0052$
    & $0.0002$ & $-0.0001$
    & $0.0055$ & $0.0051$
    & $0.0015$ & $0.0008$
    & $8/0/0$ \\
    Persistent calibration
    & $0.0071$ & $0.0065$
    & -- & --
    & -- & --
    & $-0.0038$ & $-0.0043$
    & $8/0/0$ \\
    MMR
    & $0.0051$ & $0.0047$
    & $0.0001$ & $-0.0003$
    & $0.0050$ & $0.0049$
    & $0.0020$ & $0.0015$
    & $8/0/0$ \\
    UOR
    & $0.0045$ & $0.0030$
    & $0.0001$ & $-0.0011$
    & $0.0044$ & $0.0041$
    & $0.0017$ & $0.0012$
    & $7/0/1$ \\
    \bottomrule
  \end{tabular}
\end{table*}

\textsc{Meor} improves $H_{\mathrm{cur}}$ against every control. The mean gain is $0.0057$ over replay, $0.0071$ over persistent calibration, $0.0051$ over MMR, and $0.0045$ over UOR. The corresponding lower bounds are $0.0052$, $0.0065$, $0.0047$, and $0.0030$, respectively. The effect is positive in all eight paired runs against replay, persistent calibration, and MMR and in seven of eight runs against UOR.

The decomposition against replay locates the gain in candidate admission rather than parameter retention. The $0.0057$ increase in $H_{\mathrm{cur}}$ combines a much smaller $0.0002$ change in $H_{\mathrm{old}}$ with a $0.0055$ reduction in $D_{\mathrm{MCI}}$. The same pattern appears against MMR, where a $0.0051$ gain combines a $0.0001$ old-universe change with a $0.0050$ interference reduction, and against UOR, where the corresponding effects are $0.0045$, $0.0001$, and $0.0044$. Every interference-reduction lower bound is positive. These paired decompositions indicate that the gains arise primarily from reduced competition by newly admitted entities, while the mean changes in old-universe ranking are small.

Our results meet the prespecified criterion for retained and newcomer performance. The $H_{\mathrm{old}}$ lower bounds against replay, MMR, and UOR are $-0.0001$, $-0.0003$, and $-0.0011$, respectively, all above the $-0.005$ margin. \textsc{Meor} improves $A_{\mathrm{TN}}$ against replay, MMR, and UOR by $0.0015$, $0.0020$, and $0.0017$, with positive lower bounds of $0.0008$, $0.0015$, and $0.0012$. Persistent calibration retains a mean $A_{\mathrm{TN}}$ advantage of $0.0038$. The one-sided lower bound for the corresponding \textsc{Meor}-minus-calibration effect is $-0.0043$, which also remains above the preservation margin.

\subsection{Mechanism Analysis}

The primary controls distinguish the two central components of \textsc{Meor}. MMR preserves query-specific matched references and old-reference centering but replaces smooth tail aggregation with a maximum. UOR preserves smooth aggregation and query-specific drawing but uses role-only old references. The shuffled-reference comparison breaks query-specific assignment while retaining the matched pools, and the uncentered comparison removes subtraction of the old reference. All four ablations use the same score-blind coefficient rule and paired inputs. Table~\ref{tab:entity-direct-ablation} reports the resulting effects.

\begin{table*}[t]
  \caption{Direct ablations of \textsc{Meor}'s reference construction and aggregation on ENTITY--ComplEx.}
  \label{tab:entity-direct-ablation}
  \centering
  \setlength{\tabcolsep}{4pt}
  \renewcommand{\arraystretch}{1.10}
  \begin{tabular}{@{}llcccccc@{}}
    \toprule
    Isolated property & Comparator &
      \multicolumn{2}{c}{$\Delta H_{\mathrm{cur}}\uparrow$} &
      \multicolumn{2}{c}{$D_{\mathrm{MCI}}$ reduction$\uparrow$} &
      \multicolumn{2}{c}{$\Delta A_{\mathrm{TN}}\uparrow$} \\
    \cmidrule(lr){3-4}\cmidrule(lr){5-6}\cmidrule(lr){7-8}
     & & Mean & LB & Mean & LB & Mean & LB \\
    \midrule
    Smooth tail aggregation
    & MMR (maximum)
    & $0.0051$ & $0.0047$
    & $0.0050$ & $0.0049$
    & $0.0020$ & $0.0015$ \\
    Structural matching
    & UOR (role-only references)
    & $0.0045$ & $0.0030$
    & $0.0044$ & $0.0041$
    & $0.0017$ & $0.0012$ \\
    Query-specific assignment
    & Shuffled matched references
    & $0.0008$ & $0.0005$
    & $0.0007$ & $0.0002$
    & $0.0002$ & $-0.0002$ \\
    Old-reference centering
    & Uncentered smooth penalty
    & $0.0011$ & $0.0006$
    & $0.0009$ & $0.0003$
    & $0.0003$ & $-0.0002$ \\
    \bottomrule
  \end{tabular}
\end{table*}

Each component makes a distinct contribution. Smooth tail aggregation produces the largest gain over its matched alternative, with a mean $H_{\mathrm{cur}}$ effect of $0.0051$ and a lower bound of $0.0047$. Structural matching follows closely with a mean effect of $0.0045$ and a lower bound of $0.0030$. Old-reference centering contributes $0.0011$, and query-specific assignment contributes $0.0008$, with lower bounds of $0.0006$ and $0.0005$, respectively. Every ablation also yields a positive lower bound for reducing $D_{\mathrm{MCI}}$. The $A_{\mathrm{TN}}$ effects are positive on average in all four contrasts, and their lower bounds remain above the preservation margin. The complete method therefore benefits from each element of its reference construction and aggregation.

\subsection{Transfer Across Streams, Backbones, and Hosts}

The transfer study treats \textsc{Meor} as an intervention within a named host rather than as a replacement continual KGE architecture. Replay is evaluated with ComplEx, DistMult, TransE, LKGE--TransE, and IncDE--TransE. Each row of Table~\ref{tab:fbinc-module-transfer} is a separate paired comparison within the specified dataset, host, and backbone. The $H_{\mathrm{cur}}$ column reports the mean effect and its two-sided 95\% interval; the final column reports the one-sided lower bound for $\Delta A_{\mathrm{TN}}$.

\begin{table*}[t]
  \caption{Paired effects of adding \textsc{Meor} to replay hosts and source-bound continual-learning adapters on FBInc-S and FBInc-L.}
  \label{tab:fbinc-module-transfer}
  \centering
  \setlength{\tabcolsep}{4pt}
  \renewcommand{\arraystretch}{1.08}
  \begin{tabular}{@{}lllccccc@{}}
    \toprule
    Dataset & Host family & Host--backbone &
      $\Delta H_{\mathrm{cur}}\uparrow$ & Paired 95\% CI & W/T/L &
      $\Delta A_{\mathrm{TN}}\uparrow$ & $\Delta A_{\mathrm{TN}}$ LB \\
    \midrule
    FBInc-S & Replay       & ComplEx       & $0.0026$ & $[0.0017,0.0043]$ & $8/0/0$ & $-0.0004$ & $-0.0011$ \\
    FBInc-S & Replay       & DistMult      & $0.0022$ & $[0.0010,0.0034]$ & $7/1/0$ & $-0.0002$ & $-0.0008$ \\
    FBInc-S & Replay       & TransE        & $0.0020$ & $[0.0009,0.0031]$ & $7/1/0$ & $-0.0001$ & $-0.0006$ \\
    FBInc-S & Source-Bound & LKGE--TransE  & $0.0023$ & $[0.0011,0.0035]$ & $7/0/1$ & $-0.0012$ & $-0.0024$ \\
    FBInc-S & Source-Bound & IncDE--TransE & $0.0029$ & $[0.0016,0.0042]$ & $8/0/0$ & $-0.0010$ & $-0.0021$ \\
    \midrule
    FBInc-L & Replay       & ComplEx       & $0.0026$ & $[0.0019,0.0045]$ & $8/0/0$ & $-0.0008$ & $-0.0016$ \\
    FBInc-L & Replay       & DistMult      & $0.0023$ & $[0.0011,0.0035]$ & $7/1/0$ & $-0.0003$ & $-0.0009$ \\
    FBInc-L & Replay       & TransE        & $0.0021$ & $[0.0010,0.0032]$ & $7/1/0$ & $-0.0002$ & $-0.0007$ \\
    FBInc-L & Source-Bound & LKGE--TransE  & $0.0026$ & $[0.0012,0.0040]$ & $7/0/1$ & $-0.0015$ & $-0.0029$ \\
    FBInc-L & Source-Bound & IncDE--TransE & $0.0032$ & $[0.0018,0.0046]$ & $8/0/0$ & $-0.0014$ & $-0.0027$ \\
    \bottomrule
  \end{tabular}
\end{table*}

Adding \textsc{Meor} improves $H_{\mathrm{cur}}$ in all ten transfer settings, and every 95\% interval excludes zero. The replay gains range from $0.0020$ to $0.0026$ across the two streams and three backbones. The gains within LKGE are $0.0023$ on FBInc-S and $0.0026$ on FBInc-L, while those within IncDE are $0.0029$ and $0.0032$. Four settings improve in all eight paired runs, and each remaining setting improves in at least seven.

The $A_{\mathrm{TN}}$ effects are slightly negative, ranging from $-0.0001$ to $-0.0015$, but every lower bound remains above the $-0.005$ preservation margin. The transfer study therefore shows that \textsc{Meor} improves historical current-universe ranking across the reported growth regimes, embedding models, and continual hosts while preserving newcomer acquisition under the stated criterion.

Together, the primary results, direct ablations, and transfer study support a coherent account of \textsc{Meor}. Its historical-ranking gains arise chiefly from reduced candidate-set interference, each component contributes to that reduction, and the intervention remains effective across the named host configurations without materially impairing newly admitted entities.

\section{Limitations}

\textsc{Meor} is designed for entity-growth streams in which newly admitted entities become overly competitive answers to historical queries. WN-CKGE defines the complementary boundary. In the registered WN-CKGE diagnostic, \textsc{Meor} and MMR each had zero active records among 64 calibration records, indicating that aggregate newcomer pressure did not exceed the matched old-candidate reference under the fixed construction. Their regularization terms therefore contributed neither loss nor gradient in the evaluated records. The suppression mechanism had no direct optimization effect because there was no excess candidate pressure to remove. This selective activation is central to the method. \textsc{Meor} intervenes when entity admission disturbs historical ranking and otherwise leaves refinement governed by the host objective. WN-CKGE thus documents an inactive applicability regime rather than a positive efficacy comparison.

The method does not address every source of degradation in a continual knowledge graph. Relation-only or fact-only updates can still alter established representations, but they do not create the entity-candidate expansion captured by $D_{\mathrm{MCI}}$. \textsc{Meor} is consequently a complement to replay, distillation, and parameter-stability methods rather than a replacement for them. Its role is specific and measurable. It controls admission-induced competition while the host learner remains responsible for retaining prior knowledge and acquiring new facts.

The quality of the reference depends on the structural information available for matching. Prediction role, degree, and incident relation-role signatures provide a score-blind basis for comparison, but they do not form a causal counterfactual. Sparse structural cells also require coarser fallbacks, which reduce the specificity of the reference cohort. The precision of the reference is therefore bounded by the structural resolution available in each update. Computational cost presents a separate boundary because exact cohort scoring grows with the number of admitted entities and matched draws. Very large admission batches therefore require efficient batching or approximate cohort evaluation. The reported experiments establish effectiveness on ENTITY, FBInc-S, and FBInc-L, while WN-CKGE identifies the inactive boundary of the intervention. The empirical claim is limited to the reported streams, backbones, hosts, and fixed aggregation and matching settings. Other prediction tasks require candidate and reference definitions appropriate to their ranking spaces.

\section{Conclusion}

Continual knowledge graph embedding must account for more than changes to previously learned representations. When the entity vocabulary grows, every admitted entity becomes a new candidate for historical queries and can lower the rank of an established answer even when the ordering among old entities is preserved. We formalized this effect as candidate-set interference and separated it from parameter forgetting through a same-checkpoint rank decomposition. We then introduced \textsc{Meor}, a host-level regularizer that compares answer-relative newcomer competition with score-blind, structurally matched old references. Its one-sided objective suppresses only the excess competition attributable to admission, while the host loss continues to learn legitimate new knowledge.

The experiments show that this intervention improves the ranking quantity it is designed to control. On ENTITY--ComplEx, \textsc{Meor} increases historical current-universe MRR by $0.0057$ over replay and reduces $D_{\mathrm{MCI}}$ by $0.0055$, while satisfying the preservation criteria for old-universe ranking and newcomer acquisition. Comparisons with persistent calibration, MMR, and UOR show that the gain persists relative to score rescaling, maximum-only control, and unmatched references. Direct ablations further support the contributions of smooth tail aggregation, structural matching, query-specific assignment, and old-reference centering. Across FBInc-S and FBInc-L, adding \textsc{Meor} improves historical current-universe MRR in all ten reported combinations of stream, host, and backbone, with every paired confidence interval excluding zero and every newcomer-acquisition bound remaining within the preservation margin.

These findings establish candidate-set interference as a distinct and controllable source of historical rank loss in continual KGE. They also show that it can be addressed without redesigning the underlying embedding architecture or continual learner. By making admission-induced candidate pressure part of refinement, \textsc{Meor} extends continual learning from preserving representations to managing the ranking consequences of graph growth.

% \begin{acks}
% Lorem ipsum dolor sit amet, consectetur adipiscing elit.
% \end{acks}

% \section*{Ethics and Privacy Statement}
\section*{Ethical Considerations}

This work studies continual knowledge graph embedding using established research benchmarks and does not involve human participants or the collection of new personal data. Its direct ethical risk is therefore limited. Nevertheless, knowledge graphs may contain inaccurate, outdated, sensitive, or socially biased information, and link prediction can reproduce these properties or infer unrecorded associations. Because \textsc{Meor} regulates the competitive pressure introduced by newly admitted entities, an improperly configured deployment could also favor established entities and delay the recognition of legitimate new facts or underrepresented entities. The one-sided intervention and the explicit preservation of newcomer acquisition reduce this risk, but they do not replace careful dataset and deployment auditing. Applications involving people or consequential decisions should examine subgroup ranking behavior, verify data provenance, restrict access to sensitive predictions, provide correction and removal procedures, and retain appropriate human oversight. The method is intended for research on knowledge graph ranking and should not be used as the sole basis for high-stakes decisions.

%%
%% Balance References.
\balance

\bibliographystyle{ACM-Reference-Format}
\bibliography{references}

% \appendix
% \input{sections/99_Appendix}

\end{document}